# When Exceptions are the Norm

Exploring the Role of Consent in HRI

VASANTH SARATHY, Tufts University, USA

THOMAS ARNOLD, Tufts University, USA

MATTHIAS SCHEUTZ, Tufts University, USA

HRI researchers have made major strides in developing robotic architectures that are capable of reading a limited set of social cues and producing behaviors that enhance their likeability and feeling of comfort amongst humans. However, the cues in these models are fairly direct and the interactions largely dyadic. To capture the normative qualities of interaction more robustly, we propose consent as a distinct, critical area for HRI research. Convening important insights in existing HRI work around topics like touch, proxemics, gaze, and moral norms, the notion of consent reveals key expectations that can shape how a robot acts in social space. By sorting various kinds of consent through social and legal doctrine, we delineate empirical and technical questions to meet consent challenges faced in major application domains and robotic roles. Attention to consent could show, for example, how extraordinary, norm-violating actions can be justified by agents and accepted by those around them. We argue that operationalizing ideas from legal scholarship can better guide how robotic systems might cultivate and sustain proper forms of consent.

CCS Concepts: • **Human-centered computing** → **Interaction design**.

Additional Key Words and Phrases: implicit and explicit consent, normative HRI, architectural requirements

## 1 INTRODUCTION AND MOTIVATION

Social robots are designed to engage people in an interpersonal manner, often as partners, in order to achieve positive outcomes in domains such as education, therapy, and healthcare. Their sociality also aids task-related goals in areas such as coordinated teamwork for manufacturing, search and rescue, domestic chores, and more [Breazeal et al. 2016]. HRI research has made tremendous strides in exploring and discovering the social competencies that allow for an "efficient, enjoyable, natural and meaningful relationship" [Breazeal et al. 2016]. Understanding what sorts of conduct and behavioral choices enhance or detract from positive social interactions is crucial.

In pursuit of this understanding, HRI research has organized itself along different lines. There are topics represented as types of interactions (e.g., touch based, long-term), and there are HRI subfields mapped as general aspects of interactions: proxemics [Mead and Matarić 2017; Mumm and Mutlu 2011]; approach [Walters et al. 2007]; eye-gaze and joint attention [Admoni and Scassellati 2017]; interruption [Thomaz et al. 2016]; touch [Van Erp and Toet 2013]; level of politeness [Banerjee and Chernova 2017; Briggs and Scheutz 2016; Palinko et al. 2018]. Technical approaches to equipping robotic systems to interact along such lines has generally involved computationalizing an existing social or

Authors' addresses: Vasanth Sarathy, Tufts University, 200 Boston Ave. Medford, MA, 02155, USA, vasanth.sarathy@tufts.edu; Thomas Arnold, Tufts University, 200 Boston Ave. Medford, MA, 02155, USA, thomasarnold@alumni.stanford.edu; Matthias Scheutz, Tufts University, 200 Boston Ave. Medford, MA, 02155, USA, matthias.scheutz@tufts.edu.

anthropological theory of interaction applied to the aspect being studied. HRI's empirical research helps evaluate how robots so-equipped interact in experimental settings.

Even as the field has pushed itself to more robust kinds of interactions and more sophisticated systems of planning, the prevalent notion of "social interaction" still focuses largely on dyadic exchanges. These one-on-one engagements usually rely on fairly direct cues and straightforward measurements of user preferences. For example, in approach research, it has been found that people generally prefer robots to approach them from the front and do not prefer robots to approach them from the back [Walters et al. 2007]. But there is a major "elephant in the room" in such findings, and in this paper we propose that it deserves distinct recognition as a way to organize HRI research: the notion of *consent*. Our argument is that HRI's progress in investigating interactive dynamics will not be generalizable enough without tackling the issue of consent head-on.

Consider "approach" again. There are instances when an approach from behind is acceptable and maybe even preferred, for example if a robot was pushing a person out of harm's way from an oncoming car. Wouldn't the human have consented to this act if asked and provided time? Should the robot presume consent? At first glance, it seems that that sort of reasoning is highly circumstantial and potentially too ungainly to be a cogent research problem. Like other subtleties of interaction, it could be chalked up as "context" and tackled at a future date when technology has evolved further. We argue here instead that the tools already exist to directly explore the question further and that understanding consent is central to HRI's future research.

The word "consent" typically brings to mind either notions of legality around highly deviant sexual behavior or around the appropriateness of information given to a patient about a medical procedure. However, consent's rich nuances and intricacies permeate our daily interactions in even normatively neutral environments. For example, when should a robot waiter take your plate at a restaurant? Consent can not only influence a robot's behavioral choice, but also the normative valence of a situation. Consent can turn a normatively neutral situation into a charged one, and vice versa. While consent is one factor in the larger normative fabric of society, it is its normative power and pervasiveness that provides the motivation for HRI researchers to take it seriously.

As we explicate below, the dimensions of consent can inform and flesh out existing HRI topics and findings. Returning to the subfields listed above, work on touch has already broached questions of trust, social cues, and the delicate terms on which robots should presume to touch a person [Chen et al. 2014; Van Erp and Toet 2013] . Even simple forms of robot-initiated touch evidence gender effects and workplace norms, and are evaluated accordingly [Arnold and Scheutz 2018; Wullenkord et al. 2016]. Proxemics has broached many consent-related themes, including how people compensate for robots intruding into their personal space [Sardar et al. 2012]. Spatial preferences, for example how a robot should approach someone, have received sustained attention [Koay et al. 2014; Syrdal et al. 2007], including how people adjust those preferences to aid in a robot's understanding of social signals [Mead and Mataric 2015]. The gaze of a robot can exert similar force on an interactant's social "space" and shape how a robot's role is perceived [Fiore et al. 2013; Stanton and Stevens 2017].

Sex robots have directly raised various concerns about consent, including the threat of being "ever-consenting" models of submission to abusive sexuality [Gutiu 2012]. The risk of neglecting consent as a component of interaction is to risk robots being submissive by default, allowing them to be held up as ideals against people who refuse to consent to abuse. Indeed, some wonder what consent is for such an intimate interaction [Frank and Nyholm 2017]. But it is not just a headline-grabbing issue like sex robots that makes consent salient. The socially assistive role for robots in eldercare residences, especially when those served suffer from dementia, call up numerous questions about "informed consent" and privacy among all stakeholders involved [Ienca et al. 2016; Piatt et al. 2017]. Consent may thus bind HRI even more strongly to issues for public deliberation around social institutions and their purposes.

In this paper, we introduce different kinds of consent with the help of legal theory. We then propose some concrete suggestions for how consent can be broken down and understood for settings of human-robot interaction, as well as how it can be used to drive both empirical and technical research forward. More specifically, in Section 2 we take a deep dive into five different forms of consent defined in common law, exploring, with examples, how they influence social behavior and normative expectation. In Section 3 we resurface and briefly discuss how consent issues can arise in robotic application scenarios that are relatively mundane, and not "normatively-weighty." In Section 4, we discuss the implications for the HRI community and put forth four research directions – involving research questions and architectural considerations – that take into account this nuanced understanding of consent. In Section 5, we provide some practical recommendations for how HRI researchers can make their results more consent-aware in the near-term.

## 2 LEGAL LANDSCAPE OF CONSENT UNDER THE LAW OF INTENTIONAL TORTS

### 2.1 Conventions, Norms and Laws

The normative landscape of a society can be classified into conventions, norms and laws that govern individuals' behaviors, based on the degree of their normative strength, and the level or type of sanctioning for their violations. For example, a behavior or conduct can be a convention, suggesting a stable pattern of behavior [Haynes et al. 2017], a societal regularity like people choosing to walk on the right, even when indoors. There may or may not be consequences for violating a convention. Even if they are violated, the wrongdoer is typically presented with a mild push to conform or a small dose of scoffing. Certain types of wrongful conduct can rise to the level of a norm violation. We can think of norms as conventions with an added level of sanctioning, in a stricter sense than the scoffing associated with conventions. Sanctions for norm violations can be penalizing and are intended to change the behavior of the wrongdoer. For example, walking into a Hindu temple with shoes on is a serious norm violation that can incur social sanctioning and potential banishment from the corresponding social group. Behaviors achieve their highest normative level when they are codified into law. Laws deter wrongful behaviors that society has deemed to be harmful to the basic functioning of the society. Without laws, society cannot run. Because laws are codified and interpreted repeatedly over the course of centuries, the legal landscape has attained a degree of maturity and sophistication that can serve as a guide to situate those normative elements that are lesser in strength like conventions and norms. That is, exploring and dissecting the legal landscape can provide us with some ammunition when we tackle less normatively charged situations, more common in everyday life. Below, we explore specifically the notion of *consent* from a legal standpoint, but note that the framing applies to conventions and norms as well.

We first turn to the law of *intentional torts*. Torts are a legal wrong committed upon person or property and, unlike contracts, require no prior agreement between parties. Every member of our society is obligated to damages if their actions harm others. One variety of torts, known as intentional torts, is particularly relevant and cover those situations in which intentional actions cause harm. There are several types of intentional torts, including harmful contact (battery), apprehension of harmful contact (assault), unlawful possession, use and alteration of personal property (trespass to chattels, conversion), restricting personal movement of others (false imprisonment), unlawful entry into the land of another (trespass to land), and intentional infliction of emotional distress. A crucial doctrine applicable in all types of intentional torts is the notion of consent, which can alter the morality of another's conduct [Hurd 1996] and potentially permit otherwise wrongful conduct. The doctrine is deeply rooted in history and stems from the Roman maxim *volenti non fit injuria*, latin for "to a willing person no injury is done."

Consent can convert what would be "trespass into a dinner party, a battery into a boxing match, a theft into a gift, and rape into consensual sex"[Hurd 1996]. Consent can not only convert a wrongful behavior into right, but also can grant another the right to do wrong. In this latter sense, it is less about flipping the normative valence of an act to make it right, and more about granting permission to perform what is ordinarily a morally wrong act. In both cases, consent has a tremendous amount of normative power and is an important concept that is central to every interaction. There are several forms the legal notion of consent can take and we explore each below.

## 2.2 Actual Consent

Actual consent is the "gold standard of consent" and is a person's subjective willingness to permit tortious conduct [Simons 2017]. It is generally considered to be the case that consent is limited to the person's mental state and does not also include the communicative act or signal. That a person demonstrates or evidences their consent via language, gesture and other means is separate from their truest "consent," which is their subjective acquiescence.

*A holds out her glass and asks B to take it and bring her some water. B takes the glass from A.* (1)

Here, A has consented to what would otherwise be trespass to chattels, or wrongful taking of property. Now, B's actions would need to cause some damages for them to be tortious, but we will set aside other such legal issues (damages, causation, intent etc.). Instead, we will focus on conduct that could be wrongful, and explore how consent impacts it.

We can distinguish two forms of actual consent: *express* and *inferred*. Express consent refers to situations when the consent is demonstrated explicitly either orally or in written form (i.e., linguistically). Inferred consent is when the subjective willingness is inferred from the person's conduct. (1) was an example of inferred consent, while (2) below is express.

*A says to B "You may take my plate." B takes A's plate.* (2)

The statement "you may take my plate" by A literally grants permission to B to take away the plate. A slight variant of this if A says "take my plate" leaving out the permission granting part. Note that while the statement "take my plate" does not literally state "you are permitted to take my plate", this permission can be directly inferred from the imperative form (as it does not make sense to instruct someone to take the plate without allowing them to do so – this is the crux the old deontic principle ascribed to Immanuel Kant that "ought implies can"). Utterances where the utterance semantics does not directly express the intended semantics are called "indirect speech acts" (ISAs) and are widely used in social contexts for proper conduct, often to meet politeness norms (e.g., [Briggs and Scheutz 2013]). We will discuss more about ISA's below.

Returning to express consent, one distinction that is not often drawn in the legal literature is that between express consent that is direct versus that which is indirect.

*A says to B, a bellhop, "Can you pick up my bag, please? It's quite heavy." B picks up A's bag.* (3)

Here, A was clearly consenting to B's actions, but it was certainly not direct in the sense of the semantics of the utterance. B needed to determine that A's utterance was in fact a command or at least a request for B to perform a specific action. ISAs of this sort can be highly conventionalized, and thus it is possible B interpreted the specific linguistic form just as he would the direct utterance "pick up my bag." Crucially, because consent is a mental state, we must consider the possibility that direct and indirect expressions are possible.

The degree to which a dialogue can be indirect is exemplified in (4), below.

*A arrives at an expensive restaurant and is greeted by B, the Mâitre D'. B welcomes A and then walks over close to A and says "may I?." A extends his arms. B takes off A's coat.* (4)

Here, A actually consents to B's contact and taking his property. Not only is consent signal indirect, even B's utterance soliciting the consent is indirect. In fact, B's utterance does not even reference the particular act for which B is seeking consent. Yet somehow, A has deciphered what act B wants to perform and consented to this act. The story would have been quite different if B, after asking A's permission then proceeds to do a pat-down. This would be acceptable in airport security, but not in a restaurant.

## 2.3 Apparent Consent

A second, and lesser, form of consent in which the person does not actually consent, but the actor reasonably believes that the person actually consents. In these cases, the person has not formed the subjective willingness to permit the conduct. Tort law requires not just that there be no actual consent, but also that there be no *reasonable belief* that the person consents. Unlike actual consent, which transforms or negates the wrongfulness of the act, apparent consent serves to grant permission to an act that still remains morally wrong.

*A (never been to a doctor) visits a doctor. B, the nurse, touches A to take her blood pressure. B does not reasonably know that A is averse to physical touching.* (5)

*A gets up from her table at an expensive restaurant to go to the bathroom. She places her crumpled napkin on the table by her plate. B, a waiter, comes over and folds her napkin. A has previously never been to expensive sit-down restaurants before and find this a bit strange that B touched her napkin.* (6)

In (5) (a variant of [189 1891]) and (6) B cannot rely on actual consent, because A was not actually consenting, but B might be able to use apparent consent (or possibly even presumed consent, discussed below) to preclude liability for battery in (5) and normative sanctioning in (6). A may or may not have liked her napkin to have been folded in (6) but whether she objected to it or even just did not consider it, we can be argue that she did not actually consent to it on either occasion.

The question of whether the actor can "reasonably believe" that the person consents can be based on evidence obtained from the person's conduct, others' conduct or customary norms at play.

*At the end of a successful business meeting, B reaches out and shakes A's hands before A can react. Unbeknown to B, A objects to touching and shaking hands when greeting people.* (7)

*A and B are moving about in a busy restaurant. B bumps into A as he was walking to the bathroom. A objects to touching.* (8)

*A is standing in front of B in line in a coffee shop. Believing A to be his wife based on her demeanor, pocket book she was carrying and her outfit, B gives her a surprise hug from behind. A, however, turned out not to be B's wife and objected to touching.* (9)

In (5), (7) and (8) the evidence for B's arguably reasonable belief that A was actually consenting may come from what is customary in a doctor's office or in a business meeting or in a crowded space. However, if in either case B knew that A objects to being touched, then B cannot rely on apparent consent to preclude liability. In (9), the evidence for B's reasonable belief came not from custom but directly from A's conduct and appearance.

### 2.4 Presumed Consent

We can next consider the case when the actor knows that the person did not actually consent, but reasonably believes that the person *would* actually consent, if asked. This is a counterfactual form of consent. This is different from apparent consent, where the actor reasonably believes that the person actually consents. This form of consent is typically, but not always, encountered in emergency situations when it is infeasible to obtain consent.

*A, a child, is chasing after his ball runs into oncoming traffic on a busy road. B, who's nearby, notices this and pushes A out of harm's way, but in doing so A breaks his arm.* (10)

*A is standing on the street corner. B walks up to A and taps her on the shoulder and asks her for directions.* (11)

In both (10) and (11) B is not liable to A because of presumed consent. In (10) the situation is emergent whereas in (11) it is not. Presumed consent has a high burden, additionally requiring a justification for the contact. For example, the actor must show that the contact was minor and was in fact customary in the community.

### 2.5 Constructive Consent

Often known as "implied-in-law" this form of consent is rare, but quite extreme. It considers conduct that causes socially justifiable minor acts that not only not require consent, but in fact can be performed *despite explicit objections* to the act and expressions of non-consent. Here, the standard is an objective one of whether a reasonable person would consent under the circumstances.

*A drives to work and arrives at the scene of an accident. B, a police officer, orders him to wait in the car while the scene is cleared. A demands that he be allowed to turn around and leave the scene. B refuses fearing safety risks.* (12)

B is not liable to A for false imprisonment, a variety of intentional tort, because of constructive or implied-in-law consent. This form of consent is controversial particularly because of its blatant disregard for the person's actual objections.

The concept of consent permeates our legal system beyond intentional torts. Consent is a critical defense to certain criminal conduct and plays a role in mutually beneficial contractual agreements. One interesting twist in criminal law is the existence of a category where consent is not a defense. That is, even if the person granted the actor consent, the conduct might still be deemed to be wrongful and the actor culpable. For example [198 1980; 201 2012]:

*A and B engage in a sex act in which A consents to B using a dangerous weapon during sex. B uses the weapon on A and A is harmed.* (13)

Here, B might be criminally liable in the Commonwealth of Massachusetts as consent is no defense to a charge of battery and assault with a dangerous weapon.

## 2.6 Reluctant consent

The framing of consent as a defense (to a tort claim or a criminal charge) does not consider how fervently one gives consent. If there is legal basis in evidence for one or more forms of the above consent it can be sufficient. However, exploring the amount of enthusiasm shown in regards to consent could have normative implications even if it does not have legal ones.

*A enters a taxi driven by B. B takes an unusual route that involves driving through a part of town that A associates with past emotional trauma. A says nothing.* (14)

By not objecting to this route, B might be able to rely on a subsequent defense of apparent consent when faced with an intentional infliction of emotional distress claim. It might even be the way most taxi drivers drive to the location that A intends to reach. However, A's consent is absent and, at best, reluctant. The level of A's consent might be evidenced by facial expressions suggesting it or by other gestures like looking away from the windows or closing eyes. Normatively, it seems B should take this reluctance into account and explore an alternative route. The norms at play include business aspects of ensuring that the passenger is kept happy, the human decency aspect of not having an interactant relive a past trauma and so on. Thus, the legal framework, while giving us a grounding for extreme behavioral violations, might be insufficient. We might need to consider lesser norm and convention violations that are especially evident in cases of reluctant consent.

A thorny issue in (14) is whether B had the duty to present other alternatives. The legal system is not entirely clear on this issue and some scholars have compared reluctant consent with fullness (or partial-ness) of consent and their connections to the doctrine of the assumption of risk. What risk does A assume in taking B's route? If B presented A with a second less-traumatic route, and A still reluctantly chose the first route, then the doctrine of assumption of risk may play a role in clearing B. One takeaway here is that the consenting process is much more complicated than a single interaction and typically involves a complex mix of duties and responsibilities of the actor and the mental state and behaviors of the person who arguably is deemed to have consented.

## 2.7 Other Issues

We have only scratched the surface of the intentional tort landscape around the notion of consent. There are numerous other issues around whether a person is competent or has the capacity to even grant consent. The law has established that for consent to be valid the person must have capacity to consent. Capacity is typically gauged by age and mental competence. Children, the mentally disordered, and intoxicated persons are deemed to not have the capacity to consent. This means tortious behavior will not be defensible even if consent is obtained from those without the capacity. Consent that is granted can also be revoked at any time. Consent can be conditional in that it is ineffective if there has been a failure to meet conditions that the consenter has imposed. Consent must also be voluntarily given and not coerced or obtained under duress.

Beyond intentional torts, notions of consent appear as another related legal principle in torts of negligence known as "assumption of risk." Assumption of risk is often treated as consent to conduct that is merely negligent [Simons 1987, 2006]. The doctrine of assumed risk connects consent with whether there might have been some uncertainty associated with the consequence of the conduct.

Consent is also a crucial doctrine in criminal law (as noted earlier) and in contract law. Unlike tort law, consent in contract law is centered not around internal subjective mental state, but external performance. Moreover, in contract law, consent serves as a proxy for evaluating whether the parties voluntarily entered into a contract.

Finally, it is worth clarifying that the legal principles in the paper merely serve as a guide to unpacking an otherwise complex normative landscape in many social settings. We do not require or even suggest that robots be granted any form of legal status as a person or agent. On the other hand, we want to stress that it is insufficient to treat robots as any other form of technology where consent is equivalent to user preferences. Past research has shown that robots create expectations in people that they will be less like passive technologies (e.g., washing machines and cell phones) and more like persons. While we do not take a position on the question of whether robots have "agency" in a philosophical sense, we do encourage thinking about humans in social spaces not in terms of "users" but in terms of "interactants." Robots, as part of their functions, might interact with other non-users as well. Moreover, it is not enough to express the limitations of the robot to users as users impute agency and capabilities beyond that which the robot has regardless of prior disclaimers.

## 3 CONSENT ISSUES IN ROBOT APPLICATIONS

Consent issues can arise any time there is interaction between agents. As we have seen, these consent issues come in a variety of forms and are highly dependent on the social space in which the interaction is taking place. Thus, although the notion of consent might seem somewhat dyadic, what influences the dynamics of granting and revoking of consent is not merely dyadic, but in fact influenced by the normative expectations of the entire social space. This means consent issues can arise even in scenarios that on the face are not socially or morally charged, as the social setting affects the normative expectations of the interactants. In this section, we will consider two robot applications, in increasing order of "normative-charged-ness" of scenarios in which they are typically deployed: (1) robot vacuum cleaners and (2) robot waiters.

### 3.1 Robot Vacuum Cleaners

Some have estimated that robots account for over 20% of the world's vacuum cleaners. Even if this is a bit overstated, robot vacuum cleaners are ubiquitous. These robots have a range of actuation and sensing capabilities, albeit much more limited than a humanoid robot. They have cameras, IR and laser sensors and gyros to be able to map out a house and remember what parts of the house have been cleaned and what remains. They also have other features to avoid bumps, recognize no-go areas in the house and schedule cleaning sessions. Many are beginning to come equipped with WiFi capabilities and connections to Alexa and Google Home voice-activated personal assistants. These robots are not simply passive devices that provide a service, they are mobile robots that, numerous studies have shown, cause humans to impute agency and capabilities to, beyond even that which the robot actually possesses [de Graaf 2016; Scheutz 2011; Sung et al. 2007].

The robot vacuum cleaner's primary designed function is to vacuum and clean the house, and not necessarily to interact socially with humans. Even so, before, during and after the robot performs its cleaning duties, it can and will interact with humans because it is deployed in an otherwise social space. It's owner might interact with it to decide appropriate cleaning schedules so as to take into account the presence or absence of other members of the household. The robot vacuum cleaner might inadvertently interact with non-owner humans who might be present in the space the robot is cleaning. Humans might direct the robot to go elsewhere or take an alternative cleaning route to avoid being disturbed. During its operation, the robot is also mapping the house and possibly recording video as it learns about the layout of the house and habits of the humans living there. Consent issues can arise quite easily in many of these interactions - robot getting in the way of humans moving about the house, cleaning parts of the house that are not necessarily no-go areas, but temporarily off-limits, cleaning near children's toys, some of which might be small parts, listening to private conversations in the name of listening for human commands, and recording video in private areas of the house. In many of these cases, the humans may not have actually consented, but can the robot assume there is apparent or presumed consent?

### 3.2 Robot Waiters

We have seen in the examples presented in Section 2 that consent issues can arise quite frequently in restaurant scenarios. Robot waiters constantly interact with their assigned customers, other customers walking in the restaurant, chefs, and managers. During these interactions, the robots through their appearance and functional (actuation and sensing) capabilities generate a set of expectations in the interactant humans that together establishes the social space for that particular interaction.

Distinct from robot vacuum cleaners, however, robot waiters are designed to provide a service that is inherently social, but not necessarily high-stakes or weighty. They direct customers to their seats, take orders, deliver food, check for comfort, and are meant to generally address grievances and comments issued by the customers. Each of these interactions involves entering personal spaces, touching humans, giving and taking possession of objects (some belonging to the customer). Most, if not all, of these interactions are consent-driven, in that some form of implicit or explicit consent was given or presumed, otherwise these behaviors would be problematic.

For example, consider the situation where the robot waiter is tasked with removing plates from a customer's table once they have finished eating. Customers might explicitly call out to a robot and ask for their plate to be removed. They might actually let the robot waiter know that they have finished their meal, as an indirect way to signal their desire for the plate to be removed and a check to be delivered. More often than not, customers will not make this request explicitly or even indirectly. Instead, they will provide implicit cues to suggest that they are done with a particular part of the meal and that the waiter *can* take away their plate. They issue this consent by either arranging their forks and knives a particular way, or push their plate away a bit, or even lean back away from the table as the waiter approaches. It is also possible that the customer is not really thinking about the waiter taking away their plate, but would not mind if the waiter did so. This might be conveyed via a smile that is offered after the waiter has already leaned in and grabbed the plate. The waiter may have assumed the customer was done with their meal because they had not been eating anything from it for a while. Robot waiters need to be able to read these subtle cues. Robot waiters must also be able to read cues in which the waiter does not want their plate taken away from them, even if it seems as if they are done, or they have inadvertently placed their fork and knife in a "finished" position. If the robot waiter does not pick up on refusal hand-gestures or on the customer's grabbing of a fork to show they are still eating, it can negatively impact the service and experience. It is worthwhile to note that regardless of the aesthetic appeal of the robot or of its ability to display pleasant emotions, taking away a customer's plate while they are still eating without their consent (in one of the forms mentioned earlier) can ruin the interaction.

This is just one example of a scenario where a robot's behavior, which would generally have been considered preferable, is suddenly deemed unwanted. Such a robot must be able to recognize the corresponding consent-cue to avoid the entire interaction from being impacted. Other such examples of behaviors that might require a consent-cue include: friendly touches of the customer's shoulder, filling or not filling water or wine, approaching the table to check on the customer's experience, picking up an item the customer may have dropped on the floor, clearing a path between tables by moving customer's garments, bags and chairs or requesting customers to make way, helping a customer take off their coat, and many more.

## 4 RESEARCH DIRECTIONS FOR HRI

In this section, we explore possible research directions for HRI as the legal structure presented in Section2 allows us to think about consent more deeply. But, how can this legal structure help with designing better human-robot interactions? We specifically consider four research directions or themes that the HRI research community is best positioned to address: (1) when consent is applicable, (2) how consent can be detected, (3) what happens when snap judgments must be made, and (4) what the different roles for a robot are when it is fully integrated into the normative fabric of society. For each theme, we will raise some research questions and provide architectural considerations that the HRI community might consider.

### 4.1 Applicability of Consent

When does a behavior need consenting? Presumably, there are behaviors that robots can perform that do not need consent. For example, a vacuuming cleaning robot cleaning a kitchen floor might not need consent for choosing to follow one (random) cleaning path versus another (random) cleaning path. Even in more socially-charged settings, it would be unnecessary for say an elder-care service robot to have to secure approval for every single motor movement. This would be burdensome and possibly even negatively impact the interaction itself.

Current HRI research has provided some guidance into what behaviors are preferred versus those that are unpreferred. The angle of approach example presented in Section 1 is one such contribution. We can, at a first approximation, note that unpreferred behaviors will probably need to be consented or at least need some sort of justification (viz. apparent consent, presumed consent). Also, it is possible that in many cases even preferred behaviors could require consenting. Behaviors are situated in a dynamic environment and their normative valence and the applicability of consent can, therefore, shift in time. Consider this example:

*A and B are sitting at a shared library table that can seat two. While studying, B moves some textbooks around pushing several books over to A's side of the table.* (15)

B's behavior caused an intrusion into A's space. B's behavior, however, did not start out that way. It was when B's books crossed the imaginary halfway line on the table between A and B that B needed consent. The implicit norm at play was the idea of a fair partitioning of a shared work space. The norm became activated once B's book crossed this imaginary center line.

Or consider this service example:

*A is a patient at a hospital and is recovering from a surgery in his assigned room. A receives guests and wellwishers who bring him cards, flowers and some of A's favorite take-out Chinese food, which they share during the visit. B, a janitor, is tasked with cleaning A's room and throws out not only the Chinese food containers, but also some of the cards.* (16)

Once allowed into the room, B's behavior of clearing the dirty food containers does not need consenting from A, but clearing the adjacent greeting card might. The underlying norm hinges on the agent being able to detect and consider questions of ownership, possession, and the use of space, in addition to the normative affordances offered by the objects in the environment. The greeting card serves the purpose of lifting A's spirits and therefore is an important component of the environment. B's action of clearing it might eliminate this positive utility. Note, it might be entirely inappropriate to even ask for consent to clear the greeting card as that would suggest B's normative proclivities.

*4.1.1 Research Questions.* When does a behavior turn from being consent-free to one that requires it? Are there some behaviors that require robot consent, but not human consent? Are social interaction cues of proximity, eye gaze, and so on also cues suggesting that consent may be needed? Who should be consenting in a particular situation and if that person is unavailable, should the behavior be postponed? Do humans expect robots to have the same level of understanding of consent as they do with other humans? Do they hold robots to higher standard, i.e., more conservative, when it comes to asking for consent? A popular maxim used in connection with consent in sexual acts is – "yes means yes, no means no." – do we expect the same understanding from our robots?

*4.1.2 Architectural Considerations.* One important shift in architecture design that is prompted by consent processing is that even what are otherwise considered to be automatic behaviors in robots (e.g., approach and obstacle avoidance behavior, eye gaze and blinking behavior, posture and gesturing, etc.) are subject to modulations based on consent. This, in turn, requires the architecture to have built-in suppression mechanisms that can be triggered by changing consent status and interrupt, suppress, or alter these reflex-like behaviors. In addition, the architecture requires a knowledge base with common sense knowledge about the different consenting conditions and the appropriate modulations. For example, it will require general knowledge about possession of property, what actions are and are not allowed with another person's property, and under what circumstances the robot can dispense with certain rules and norms because there are role-based expectations that trump them. In the above example, even though the food container was technically owned by patient A, a janitor robot B would have had to make the inference that based on its role to clean up used items – and an empty food container was a used item – the ownership of the container was no longer an issue and it was, in fact, expected from B to take the container away from A. This is a case where two norms clash, the norm of following ownership rules versus the norm of fulfilling one's duties based on one's roles. The challenge then is to detect that the role-based norm of cleaning up the container trumps the ownership-based norm of not touching it precisely because it implies consent by the owner to do so.

Detecting shifting consent-dependencies of behaviors is particularly challenging from a technical standpoint. Detecting consent applicability shares many similar technical aspects to detecting consent itself. The robot will need to detect a variety of cues from its interactant, others in the environment, and other animate and inanimate objects that constitute the social space. A technical designer must consider what level of symbolic abstraction does a behavior need to be represented as before it qualifies as one applicable for consent. Clearly, step-wise motor movements and joint angles might be too specific and not necessarily applicable. On the other hand, higher-level behaviors of swinging arms and even higher-level conduct of picking up and manipulating objects might be candidates for consent-dependency. It is also possible that the robot might need to consider the nature of objects it is handling (e.g., knives versus a bouquet of roses) when performing certain behaviors.

## 4.2 Detecting Consent

At the core of consent-related considerations that we have discussed in this paper is the question of how and what consent cues are detected. At first glance it seems like language, gestures, facial expressions, eye gaze, approach, proximity can all serve as valid vehicles to communicate one's subjective intent of actual consent or objection. But, how can a robot detect consent expressed through an ISA, or through the usage of objects in the environment? Consider this example:

*A was sitting at the bar getting a drink that he already paid for. A drinks most, but not all, of the drink and then leaves the bar. B a waiter did not see A leave. B takes away A's almost empty mug.* (17)

Presumably, B's conduct was normatively appropriate given that A might not have objected to the mug being taken away. Note, the mug actually belongs to the restaurant, so there is no legal consequence here. We nevertheless use the legal framework to inform our discussion about consent norms and consent conventions. So, even if A had just stepped out for a smoke or to take a call, B's actions might have been justified. Now, contrast this example with one where A leaves his bag on the seat and goes to the restroom. Here, B's actions of taking away the mug might not be justified. A might object that he had not finished his drink and his intentions were made clear by him leaving his bag on the chair.

There are many different types of consent cues. Some cues are *consent-granting*, in which case they have an illocutionary force of conveying permission to a behavior. Alternatively, some cues are *consent-denying*, in which case they have the force of conveying explicit non-consent to a behavior. In some instances, the person is not yet ready to signal, but they will eventually. These *consent-withholding* cues suggest that the actor return at a later specified or unspecified point to inquire about the consent. For example,

*B, a waiter, is about refill A's wine glass. A gestures with a downward facing open palm and says, "not yet." B retracts.* (18)

Here, A did not say to never fill his glass, instead withheld consent until some later point. Now, there is a subtlety here where A might have instead just said "no", which might have been a consent-denying cue, or if prior consent had been granted, a *consent-revoking* cue, taking away a prior consent. Finally, the consent itself might be conditioned and expressed via a *consent-conditional* cue of A saying "only if it is a 1991 Argentinian Malbec" suggesting that refilling is only acceptable for a particular variety of wine.

These different types of cues not only convey whether or not consent is given, but also specify which behavior they refer to. In (18), when A says "not yet" they are not withholding eventual consent for B pouring the wine on them. The content of the consent is difficult question that poses many challenges.

The legal landscape also suggests that for succeeding on a defense of consent, the actor must also prove that the person has the capacity to consent, is consenting voluntarily and has all the information necessary to decide. The latter of these elements is the basis for the doctrine of informed consent, common in medical settings.

*4.2.1 Research Questions.* How to detect social and environmental consent cues? Are there certain cues that transcend situations and universally signify consent or non-consent for any behavior? Is it enough for the robot to detect and then immediately act on a consent cue or should it ask for confirmation? What is the role of uncertainty in detection?

Crucially, what if the cue was not only unclear in a noisy environment, but also ambiguous in its scope and content? If the robot is unable to glean actual consent, should it explicitly ask the human or should it be satisfied with inferring apparent consent?

*4.2.2 Architectural Considerations.* To be able to detect cues, the robotic architecture must be able to first process and integrate percepts that can serve as cues obtained through various sensory modalities. Consider linguistic cues, which, although they seem like the clearest type of cue, can still be a technical challenge to process. Here, processing consent involves a degree of intent recognition and processing of potentially indirect cues (via indirect speech acts). Architecturally, it might not be enough to implement statistical natural language processing and speech recognition systems that do not have a way to represent contextual variations in the illocutionary and perlocutionary force of speech acts. A way around this issue would be to explicitly design an architecture that allows for performing cue-based inference. An approach might be to use a rule-based strategy where implication rules can represent consent-cues as antecedents and behaviors as consequents. Cues then trigger rules that determine appropriate behavior. The normative propriety of the behavior can then computed via logical modus ponens based on the perception of the relevant consent cues. A variant of this rule-based approach can be seen in indirect speech research ([Briggs and Scheutz 2013]). This sort of architecture and knowledge representation has also been shown to be amenable to non-statistical, instruction based teaching and learning via natural language.

In addition to linguistic cues, the robot may have to perform inference from visual percepts as in (17). In such cases, the robotic architecture must be equipped to not only recognize objects, but also generate more holistic representations of a scene. This means object properties need to be inferred and, in dynamic environments, objects and their properties will need to be tracked. These object representations will then need to be integrated with other modalities including the linguistic cues noted above in order to generate a coherent *story* for whether or not the robot has deemed there to be consent in a particular situation.

But that only covers cases of actual consent. The cues in actual consent are typically limited to those that result from the conduct of the person consenting. However, in apparent consent situations, the cues can come from not only from the person who is the target of the behavior in question, but also from others and from customary aspects of the situation.

The robot must also be able to understand the interactants' capacity to consent and ensure that they are provided with all the information they need to make their consent decision. Robots working with certain vulnerable populations – children, elderly, substance-abusers, mentally disordered – must account for these aspects.

## 4.3 When Consent Does Not Matter

What the legal landscape has shown us is that there are situations where an otherwise unconsented behavior is appropriate (presumed consent and constructive consent) and consented behavior is still inappropriate (consent not a defense to certain criminal charges).

In situations like emergencies where consent is presumed, the law places a high burden on those looking to use this principle in an affirmative defense to an intentional tort. Presumed consent falls squarely below actual and apparent consent and has limited scope. But, these are situations in which our robots might find themselves, possibly even more frequently than imagined. Search-and-rescue robots, space robots, medical robots and robots in many emergent situations must take quick action in order to prevent a larger harm.

In certain cases, sex robots engaged in BDSM activities might have received consent for certain acts, but the law has held that performing these acts (particularly those with dangerous weapons) can still remain a criminal charge. On a less egregious note, sex robots that bring spontaneity and creativity into a sexual encounter (e.g., a surprise tickle) might need to rely on presumed consent to provide a positive social interaction.

*4.3.1 Research Questions.* Should a robot ever presume consent? If a robot does presume consent, should its scope be kept limited to a few specifically designed behaviors. Presumed consent is particularly controversial and it, combined with constructive consent, start tearing away at personal autonomy. If a person's consenting of a behavior is taken for granted or presumed given or worse, ignored, is there not a taking away of a moral right to autonomy? Should we give robots this privilege? What if it ends up helping us? Moreover, do we humans expect this from our rescue robots? Some in the HRI community have explored the role of supererogatory actions (like in (10)) and whether we expect a street cleaning robot to go out of its way to save a child from an oncoming car. The technical challenges for designing such a robot share many aspects of cue detection and norm representations noted earlier. In cases where consent is not a defense, should robots be designed to never perform these behaviors? How can we balance robot creativity, spontaneity and risk-taking behaviors (all of which might have positive social value) with normative propriety? Can the robot presume consent if it enhances a social experience?

*4.3.2 Architectural Considerations.* While the question of whether or not consent matters in certain special situations is largely an empirical and philosophical one, we can still evaluate what architectural capacities a robot must have. Uniquely, robots must be equipped with suitable architectural components to ensure that it does not become a passerby, or bystander, during critical situations. Thus, it is important for there to be coordination within the architecture for integrating cues suggesting an emergent situation and the ability to reason against norm expectations. Crucially, such reasoning is required as the robot needs to deduce that a norm violation might be necessary in order to achieve a greater good. The architecture will need mechanisms to reason with multiple such mutually inconsistent and conflicting norms, all in real-time under extreme time-pressures imposed by the emergent situation.

## 4.4 Robot Roles

Thus far, we have looked at several examples in which we examined the propriety of actions of B in interactions with A. While it is natural to think about the role of a robot as B, and the human as A, in reality, the role of the robot could be quite varied. Here, we can explore some of the alternative roles the robot plays in an interaction.

*4.4.1 Research Questions.* First, we can consider the robot as a consenter (or protester), i.e., A. Robots in this role raises many questions about robot rights and autonomy. Should the robot be able to consent at all? A narrow view of robot rights would suggest that robot consent does not matter. This view, in turn, might increase the scope of presumed consent or constructive consent in robot interactions. Can we then abuse a robot?

*B verbally abuses A a personal assistant robot by shouting expletives.* (19)

Some might argue that A's consent towards this behavior is irrelevant as A is a machine. However, the counter argument to this view is that B's verbal abuse sets the wrong normative tone in the environment and therefore should be seen as wrongful. There may not be emotional distress for A, but there could be for another, say C, who is watching this interaction.

We can bring in another interactant C who might be a third-party to the interaction. C could play the role of sanctioner and function to right normative wrongs in an environment. For example, C could be a manager of a restaurant in which B a waiter is serving A a customer. Alternatively, we can think of C as a sanctioner role that could be played by A themselves.

In such a role, we must ask if it is okay for the robot to call-out and sanction normatively harmful behavior (see [Fessler 2018] for Alexa pushing back against abuse). In (19) could the robot defend itself and protest A's abuse? Should the robot do such a thing?

Another role a robot could take is not that of a moral arbiter, but that of the elusive "reasonable person". In much of the legal scholarship one aspect that has been at the center of many debates is the the objective standard of a "reasonable person", one who exercises average skill and judgment and can objectively assess a situation and arrive at the correct solution. Many legal scholars have argued that such a person does not exist and is, in fact, by no means average. Nevertheless, robots might present themselves in the unique position to serve as such a reasonable person. Can these reasonable robots help resolve conflicts by performing an unbiased and objective analysis of the facts and situation?

*4.4.2 Architectural Considerations.* The architectural capabilities we have discussed thus far apply to the robots that serve as sanctioners or reasonable persons. Of particular note, these robots must be able to detect norm violations, and so also have architectural components to compare normative cues with norm-based expectations to detect norm violations. Robots in these varying roles also have the function of deciding if the justifications provided by parties is acceptable and as such must have mechanisms to infer consent by exploring the purposes of actions. Investigations into human-robot teamwork have even included their possible role as repairers of social conflict [Jung et al. 2015]. Robots must also maintain honesty [Hoffman et al. 2015] and sometimes even keep secrets [Kahn Jr et al. 2015] in order to fulfill the responsibilities [Asaro 2007] imposed by their roles. This also means that robotic architectures serving any role must be able to represent various supporting aspects of the situations – ownership, possession, resulting permissions and obligations and whether or not agents have the capacity to consent and have done so voluntarily with the needed information. The architectures must also be able to determine precedence relationships to help enable consent inference.

## 5 NEAR-TERM NEXT STEPS FOR HRI

In the previous section, we discussed four research directions (or themes) that are opened for the HRI community by the more nuanced consideration of consent. Critically, we proposed research questions and technical (architectural) design issues that will need to be addressed if we are to design fully consent-aware robots. In some sense, we presented a research vision for normative HRI that is focused on much-needed technical advancements that are likely to be longer-term. But, what can HRI researchers start doing today? We do not yet have all the technical capabilities to build the architectural components needed for fully consensual agents. However, as we have discussed in Section 3, robots are already deployed in many real world social settings and will need to be able to handle at least some partial set of consent cues. How can HRI researchers design and evaluate more consent-aware interactions with these current robots? In this section, we provide some recommendations for how current HRI researchers can adapt their design and evaluation methodologies to begin accounting for consent. We expect that by doing so, this will guide future technical developments in the right direction.

### 5.1 Early-Stage Interaction Design

HRI design is a truly collaborative endeavor and tackling some of the consent issues might need to go beyond architectural modifications of robots. That is, it is insufficient to only take a robot/technical-centric view and place all the burden of consent-management to the robot and its underlying A.I. Accounting for consent must begin at the earliest stages of interaction design. Interaction design involves considering not just behaviors in isolation, but the generation of sequences of behaviors and how such sequences influence and are influenced by particular application scenarios. Behaviors could include both dialogue and non-dialogue (body movements), and interaction forms the overall specification of the set of behaviors that technical researchers must build into their robots. Interaction designers can proactively consider consent when designing specific interactions by asking the research questions we presented above *early* in the design stages of storyboarding and prototyping. Asking when and how consent will apply to particular scenarios by designers at this stage will likely prove crucial when designing the specifics of the interaction itself and contain the set of robot capabilities. As behaviors are evaluated against a scenario and flows constructed, interaction designers can asking critically *when* and *why* certain behaviors are acceptable or not, clueing them into whether or not consent applies in a particular situation. For example, asking when and why it is okay for a robot waiter to fill an empty water glass raises many interesting questions about the situation and the nature of the social space that might need to be answered before a suitable interaction can be designed. The answers might suggest entirely different interactions for an expensive restaurant versus a fast food restaurant. A deeper inquiry at that point might shed some light on what factors might turn acceptable behaviors into unacceptable ones and vice versa. For example, customers in expensive restaurant might provide consent cues that are much more indirect and implicit than those at fast food restaurants. Similarly, the interaction designer could attempt to model the mental state of users in the scenario to consider what these users would consider reasonable deviations from acceptable behavior. If the scenario involves emergencies, interaction designers will be able to directly incorporate notions of presumed consent. For example, when a robot detects that a human is about to get run over, it can assume presumed consent for pushing the person out of harms way, even though pushing normally would violate a tort.

### 5.2 Experimental Evaluation and User Studies

When robotic systems are introduced into a real world application scenario, they can have a significant impact on usability, user experience, and social aspects like impact and acceptance. Prior to such an introduction, a strong evaluation methodology is needed to enable fair comparisons of competing HRI systems along various dimensions including efficiency, feasibility, safety, and social and psychological aspects. There have been many approaches proposed in the literature for the most suitable set of metrics for evaluating robots. A prominent one is by Steinfeld et al. [Steinfeld et al. 2006] who propose a set of task-specific as well as common metrics for assessing a robot's performance. In this comprehensive work, they discuss some socially-driven metrics (persuasiveness, trust, engagement and compliance), but these are largely discussed in connection with robots whose primary function is to interact socially with people. As we have discussed, socially-relevant consent issues can arise even when the robot's primary function is not social interaction; consent issues can be evoked from the robot simply operating in a social space. The common metrics proposed by Steinfeld et al. do not consider such normative and socio-cultural aspects of a scenario in which the robot is deployed. Thus, urgent future work is needed to develop suitable metrics and measurement instruments that explicitly probe for consent.

In addition, we propose that interaction researchers studying behaviors, but not necessarily studying consent, can and should provide some more guidance about the ecological validity of their findings. They can do this by asking when preferred behaviors become dispreferred and vice-versa. User-study questionnaires already ask questions about user preferences and likeability of behaviors. These can be extended to include questions about dislikeability and what makes a certain behavior inappropriate, and when dispreferred behavior could become preferred. By exploring ways to extend these questionnaires we can study how consent influences existing interaction research findings and provide some new ideas for future research inquiries.

More generally, when robot behaviors are being evaluated for a particular use-case, HRI researchers must consider the nature of normative expectations held by humans present in that use-case, and ensure that their experimental subject-population is representative of those in the real world. Most HRI user studies obtain demographic information either to directly study gender or some other effect relating to the identity of participants, or indirectly to correct for individual differences across a population. We are suggesting here that consent-preferences are an equally important (an possibly biasing or confounding) factor to either study or at least correct for in user studies.

Finally, when drawing conclusions from user-studies HRI researchers must make explicit the types of consent-related aspects they have accounted for in their experiment. That is, their conclusions and claims about behavioral preferences must be accompanied by the scope and nature of the scenario(s) in which those behaviors are situated. During the peer-review process, we urge reviewers to extend this inquiry and consider how consent or lack thereof can impact these results and push the researchers to make explicit these limits.

### 5.3 Revisiting Past HRI Experiments

The future outlook for HRI research incorporating this rich notion of consent is quite exciting, and we hope the suggestions provided in this paper will facilitate a smoother transfer of future interaction research results into real world robotic application domains. That said, it is still unclear what we are to do with past and current research results like, for example, research on "robot approach" that we alluded to in Section 1. A formidable challenge lies in understanding the scope of the results in not only approach research, but also dialogue, eye gaze, proxemics, touch and other interactionally-relevant behaviors. We know that behavioral preferences can be highly contextualized and dictated by the task at hand. But, it is unclear how the results for particular studies can be generalized beyond the specific experimental setting in which they were designed.

For example, in eye gaze research, HRI researchers have suggested that high levels of mutual gaze express feelings of trust and extroversion, and gaze aversions express feelings of distrust and introversion [Admoni and Scassellati 2017]. However, this is not universally true, as an unwelcome extended eye gaze might be suspect and can adversely impact trust. In conversational contexts, eye contact is seen as more acceptable in emotionally neutral topics but not so when the conversation is embarrassing [Admoni and Scassellati 2017], suggesting that the topic of conversation could be a modulating factor when evaluating robot perceptions. Although there could be other modulating factors in particular scenarios, we propose that consent is a key modulating factor in all HRI settings, and whether or not certain eye gaze strategies are deemed socially acceptable (or any other socially relevant evaluation metric) is dependent on whether the interactants mutually consented to each others behavioral choices. In situations where there is no explicit consent, the question then becomes if the social setting provides a reasonable basis for establishing some implicit form of consent. A social space that places a higher bar on consent, will not allow for prolonged eye contact even if conversations are emotionally neutral, without a reasonable basis for it. In order to extend the applicability of many such HRI studies into *real* interactional contexts, what we need is a deeper understanding of the social spaces anticipated by these studies.

We need a better sense for what sorts of factors need to be present or absent to reasonably establish that either consent can be apparent or presumed in the social space, or actual consent was indeed given.

As another example, consider research in politeness in natural language. HRI researchers first looked at speech modifications as a way to modulate politeness and thereby improve human perceptions of robot helpers. Torry et al. [Torrey et al. 2013] argued that robot helpers create a positive impression when using hedges ("I guess" or "kind of" or "probably") and discourse markers ("I mean" or "like you know"). Their study involved humans observing robots employing those dialogue strategies, not humans directly interacting with such robots. Strait et al. [Strait et al. 2014] subsequently noted that such speech modification strategies alone are insufficient and other aspects of the interactional scenario like participants' personality, dialogue efficiency and task success also impacted the perceptions of the robot. They explored the relative effects of communication strategy as modulated by interaction modality and presence as well as robot appearance. Their results showed that some prior claims made by Torry et al. do not generalize to real interactional scenarios where other modulatory factors might be at play. Their conclusions leave open the possibilities for further situational factors that might modulate interactions. We believe this is a step in the right direction, and we urge HRI researcher to make more explicit the spectrum of modulatory factors that might be present in a particular HRI scenario. Specifically though, we reiterate that consent is one such modulatory factor, and an extremely important one at that. We believe the effect of consent is critical to evaluating robot perceptions, and on par with other influential factors such as gender effects.

Regardless of where one stands on the relative importance of the role of consent over other situational factors, it is fair to say that since consent can significantly impact all HRI interactions, it must be explicitly accounted for in our HRI experiments. One direction might be to redo some of the more significant past experiments to include some explicit consent measures and possibly even using consent/no-consent as an independent variable in the design. So, the robot-approach results could be redesigned by taking the presupposed consent structure of the context in which they are performed into account. On the one hand this might sound like bad news, but we think it is not so and will actually substantially strengthen those results that survive this repetition, and provide new opportunities for those results that show some variation when corrected for consent. Importantly, such a revisiting of past experiments might be necessary in order to be able to understand what those interaction design results actually mean when those interaction designs are translated into real world applications deployed in social spaces requiring consent-awareness. We might need to go back to the drawing board on some experiments to get additional results for cases not previously considered or conflated due to the lack of a consent dimension in the experimental design.

## 6   CONCLUSION

In this paper we have presented consent as an crucial area for HRI research. Consent often defines human-robot interactions for their entire duration, rather than being just a precondition that is established or refused beforehand. We have outlined levels of consent to show how subtle social cues can change the nature of what a social robots roles are and what actions they can do. From several exemplary scenarios, it is clear that the HRI community can extend insights of existing research into richer areas of human-robot interface. Our brief architectural considerations suggest how robotic systems, designed with architectures incorporating rules-based and context-aware elements, could best recognize how consent is sought, maintained, and respected.

As robots enter into the rough terrain of social spaces, consent will unavoidably take on cultural and political valences. Even seen as tools, robot will assume representations of broader values, for better or worse. Societal critiques have stressed the harm of gendered norms of silence and disregard for consent, and steering clear of consent still implicates robotic systems in how they are embodied. This reflects why for broader discussions of AI ethics, including law and policy, HRI research can be a strong voice for otherwise neglected risks and opportunities posed by embodied automation. It can be at the forefront of guiding how human society can more carefully consent to the technologies meant to serve it.